\documentclass{article}

\usepackage{arxiv}

\usepackage[utf8]{inputenc} % allow utf-8 input
\usepackage[T1]{fontenc}    % use 8-bit T1 fonts
\usepackage{hyperref}       % hyperlinks
\usepackage{url}            % simple URL typesetting
\usepackage{booktabs}       % professional-quality tables
\usepackage{amsfonts}       % blackboard math symbols
\usepackage{nicefrac}       % compact symbols for 1/2, etc.
\usepackage{microtype}      % microtypography
\usepackage{multirow}
\usepackage{lipsum}
\usepackage{graphicx}
\usepackage{amssymb}
\usepackage{makecell}
\usepackage{subcaption}
\usepackage{bm}
\usepackage{float}
\usepackage{amsmath}
\usepackage{pbox}
\usepackage{authblk}

\title{Augmenting Variational Autoencoders with Sparse Labels: \\
A unified framework for unsupervised, semi-(un)supervised, and supervised learning}
\author[1]{Felix Berkhahn \thanks{felix.berkhahn@relayr.io}}
\author[1]{Richard Keys \thanks{richard.keys@relayr.io}}
\author[1]{Wajih Ouertani \thanks{wajih.ouertani@relayr.io}}
\author[1]{Nikhil Shetty \thanks{nikhil.shetty@relayr.io}}
\author[1]{Dominik Gei\ss ler \thanks{dominik.geissler@relayr.io}}
\affil[1]{Relayr (GmbH), Munich}

\begin{document}
\maketitle

\begin{abstract}
We present a new flavor of Variational Autoencoder (VAE) that interpolates seamlessly between unsupervised, semi-supervised and fully supervised learning domains. We show that unlabeled datapoints not only boost unsupervised tasks, but also the classification performance. Vice versa, every label not only improves classification, but also unsupervised tasks. The proposed architecture is simple: A classification layer is connected to the topmost encoder layer, and then combined with the resampled latent layer for the decoder. The usual evidence lower bound (ELBO) loss is supplemented with a supervised loss target on this classification layer that is only applied for labeled datapoints.  This simplicity allows for extending any existing VAE model to our proposed semi-supervised framework with minimal effort. In the context of classification, we found that this approach even outperforms a direct supervised setup. 
\end{abstract}


\section{Introduction}
In many domains, unlabeled data is abundant, whereas obtaining rich labels may be time consuming, expensive and rely on manual annotations. As such, the value proposition of semi-supervised learning algorithms is immense: It allows us to train well performing predictive systems with only a fraction of labeled datapoints.

In this paper, we present a new flavor of Variational Autoencoder (VAE) that enables semi-supervised learning. The model architecture requires only minimal modifications on any given purely unsupervised VAE. The semi-supervised classification accuracy has similar performance as slightly more complex approaches known in the literature \cite{kingma2014semisupervised, kossyk2019}. This was benchmarked using the MNIST (section \ref{sec:mnist:classification}), Fashion-MNIST (section \ref{sec:fashionmnist:classification}) and UCI-HAR (section \ref{sec:ucihar}) data sets. We verified that even if every single datapoint is labeled, framing the training process in the context of VAE training improves the classification accuracy compared to the common way of training the classification network in isolation. We conjecture that supplementing the classification loss with the VAE loss forces the network to learn better representations of the data. Here the VAE reconstruction task acts as a regularizer during training of the classification network.

We also verified that the availability of labels helps the model to find better latent representations of the data: We used the betaVAE disentanglement metric to asses the quality of the found representations (section \ref{sec:disentanglment}). Furthermore, we applied the VAEs to the problem of anomaly detection, and observed that its performance increases when the model is trained with additional labeled samples - see sections \ref{sec:mnist:anomaly}, \ref{sec:fashionmnist:anomaly} and \ref{sec:ucihar} for benchmarks on MNIST, Fashion-MNIST and UCI-HAR respectively. In that sense, not only is the reconstruction of the model boosted by the availability of unlabeled datapoints (which is the normal semi-supervised setup), but vice versa the anomaly detection performance is also improved by the availability of labels.

In summary, we have developed a model which adapts seamlessly on the full 0-100\% range of available labels. The result is a ‘unified’ model in which the anomaly detection capability is improved by any available label, and vice versa in which the predictive capability is significantly boosted by the abundance of unlabeled data. This paper provides a more thorough investigation and benchmark of the concepts which were published in a blog post in 2018 \cite{berkhahn2018onemodelblogpost}.

\section{Model}
\subsection{Model architecture}
\label{sec:model_architecture}
\begin{figure*}[h]
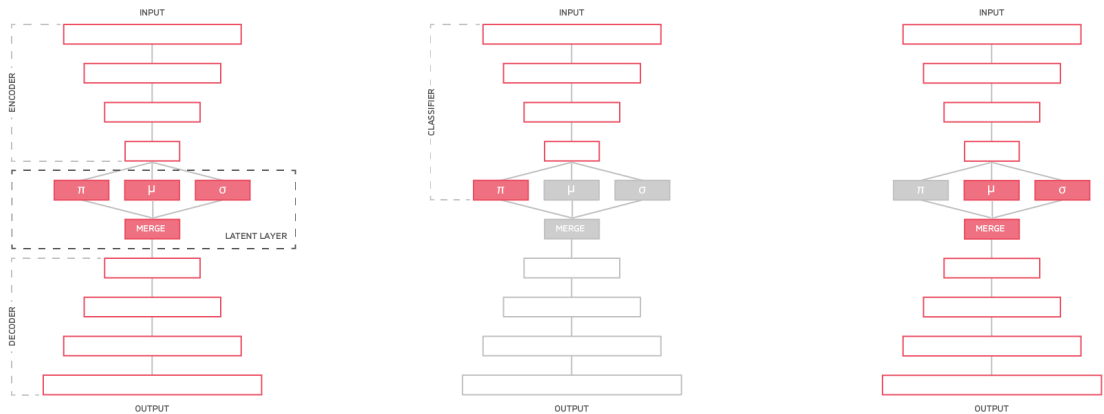

    \centering
    \begin{subfigure}[t]{0.33\linewidth}
        \centering
        \includegraphics[width=\linewidth]{model_ss.png}
	   \caption{Semi-Supervised VAE, model $SS$}
        \label{fig:model_architecture:ssvae}
    \end{subfigure}%
    ~
    \begin{subfigure}[t]{0.33\linewidth}
        \centering
        \includegraphics[width=\linewidth]{model_es.png}
        \caption{Supervised Classifier, model $ES$}
        \label{fig:model_architecture:es}
    \end{subfigure}%
    ~
    \begin{subfigure}[t]{0.33\linewidth}
        \centering
        \includegraphics[width=\linewidth]{model_eu.png}
        \caption{Unsupervised Anomaly Detector, model $EU$}
        \label{fig:model_architecture:eu}
    \end{subfigure}%
    \caption{Comparison of our model architecture (a) to the supervised (b) / unsupervised (c) equivalents. The greyed out cells shown in (b) and (c) are not part of the models, and highlight the difference to our model (a). Note that the $\pi$ layer (and its loss) represents the extension to the standard VAE proposed in this paper.}
    \label{fig:model_architectures}
\end{figure*}

The general model architecture is depicted in Figure \ref{fig:model_architecture:ssvae}. As can be seen, the model is an extension of the original VAE \cite{kingma2013vae} which is depicted in  Figure \ref{fig:model_architecture:eu}. The only addition is that a classification layer $\pi$ (typically a one-hot classifying layer using softmax activation) is introduced that is attached to the topmost encoder layer. The $\mu$ and $\sigma$ layer encode the mean and standard deviation of the gaussian prior in the latent layer: 
\begin{equation}
\label{eq:latent_normal}
p(\textbf{z}) = \mathcal{N}(\textbf{z} | \mu, \sigma)
\end{equation}

After sampling the latent variable $\textbf{z}$ using the probability distribution (\ref{eq:latent_normal}), $\textbf{z}$ and the activations of $\pi$ are merged and fed into the decoder $p_{\theta}$:
\begin{equation}
\textbf{x}_{recon} \sim p_{\theta}(\pi, \textbf{z}) = p_{\theta}(\pi \oplus \textbf{z})
\end{equation}
where $\textbf{x}_{recon}$ denotes the reconstructed data of the decoder. Hence, the classification predictions are also contributing to the reconstruction of the data.

\subsection{Loss function}
\label{sec:loss_function}
We propose an ad-hoc modification of the standard VAE evidence lower bound $L_{ELBO}$ \cite{kingma2013vae} loss function:
\begin{equation}
\label{eq:loss}
L = L_{ELBO} + L_{cl}
\end{equation}
where
\begin{eqnarray}
L_{ELBO} &=& \mathbb{E}_{z \sim p(z)}\left[ \log \left(p_{\theta}(\textbf{x}|y, \textbf{z}) \right) \right] - KL\left[ p_{\phi}(z) || \mathcal{N}(0, \textbf{I}) \right] \\ \label{eq:elbo}
L_{cl} &=& - \frac{\alpha(y)}{\text{\#labeled}} \sum_i  y_i \cdot \log(\pi_i) \label{eq:cl_loss}
\end{eqnarray}
$KL\left[ p_{\phi}(z) || \mathcal{N}(0, \textbf{I}) \right] $ denotes the Kullback-Leibler divergence of $p_{\phi}(z)$ and a standard normal distribution, which is only applied to the latent variables $\textbf{z}$, but not the labels $\pi$. $p_{\phi}(z)$ is the probability distribution of the latent variable generated by the encoder. $y$ represents the label of the datapoint. $\alpha(y)$ is equal to zero if there is no label (ie $y$ belongs to the 'unlabeled class'), else it is one. Normalizing $\alpha$ to the number of all labeled datapoints per batch aids stabilizing training. $L_{cl}$ denotes the classification log-loss.

\subsection{Upsampling the labeled data}
\label{sec:upsampling}
In order to prevent artificial noise from a stochastic number of labeled contributions in the log-loss term (\ref{eq:cl_loss}), we chose to not only normalize this term but also fix the number of labeled samples per batch: Besides the completely (un)supervised edge cases, we sampled datasets such that each batch contained labeled and unlabeled samples in a ratio of 1:1. Additionally, this prevents unlabeled datapoints from dominating training in cases of very sparsely labeled datasets. 

\subsection{Differences to Kingma's VAE \cite{kingma2014semisupervised}}
\label{sec:differences}
Our work is largely inspired by \cite{kingma2014semisupervised}. However, the model we are proposing differs from the model M2 of \cite{kingma2014semisupervised} in several aspects as shown in Table \ref{tab:differences}.

\begin{table}[h]
\begin{center}
\caption{Differences to Kingma et al.}
\label{tab:differences}
\begin{tabular}{ l | c | c}
   & our model & Kingma et al \\
   \hline
  encoder & \makecell{single encoder network \\ sharing weights $q_{\phi}(\textbf{z}, y|\textbf{x})$} & \makecell{two independent encoder networks \\ $q_{\phi}(\textbf{z}, y|\textbf{x}) = q_{\phi}(\textbf{z}|\textbf{x}) q_{\phi}(y|\textbf{x}) $ } \\ 
  \hline
  latent layer & \makecell{latent activations only \\ depend on $\textbf{x}$: $\mu(\textbf{x})$ } & \makecell{latent activations depend on \\ both \textbf{x} and $y$: $\mu(\textbf{x}, y)$}  \\
\hline
treatment of unlabeled data & \makecell{$\alpha(y)$ will omit \\ contribution to $L_{cl}$ } & unknown $y$ is summed over \\
\end{tabular}
\end{center}
\end{table}

The simplicity of our model allows to turn any existent VAE into a semi-supervised VAE by simply adding the $\pi$ layer and extending the loss function. In particular, all learned weights can directly be reused when transitioning into the semi-supervised learning scenario. This is very useful, as in many real world applications, a labeled dataset (even partially labeled) is only built up over time and not available at project initiation. 

\subsection{Classification - Decoder as a regularizer}
\label{sec:decoder_regularizer}
We benchmarked the classification performance of our model for various data sets (MNIST, Fashion-MNIST, UCI-HAR, see sections \ref{sec:fashionmnist:classification}, \ref{sec:fashionmnist:classification} and \ref{sec:ucihar}) as a function of available labels. Not surprisingly, more labeled or unlabeled samples generally improves performance.

Moreover, we also tested our model in the scenario where \textit{all} datapoints were labeled. Interestingly, we found that the obtained model was performing better than training the same classification model (Figure \ref{fig:model_architecture:es}) in a standard supervised scenario. In other words, framing the training process in the context of VAE training allows the classification network to learn better weights compared to training it the 'standard way' with only the $\pi$ classification loss. 

The additional training target of reproducing the input via the decoder forces the network to learn more meaningful representations in its deeper layers, from which the classification benefits. The decoder and VAE training act as a regularizer, as it challenges the network to find more subtle and granular representations of the input data, i.e. it will combat overfitting. At the same time, these representations are meaningful, as they contain valuable information about how to reconstruct the datapoint properly - hence it is expected that they enhance any task built on top of them (for instance, classification). 

\subsection{Semi-Unsupervised learning}
As we have seen in the previous sections, the availability of unlabeled datapoints aids the model to form better representations in its deeper layers, hence enabling semi-supervised learning. Maybe the opposite is true as well: Does the availability of labels also aid with finding better representations? Does it perform better on reconstruction related tasks such as anomaly detection?

This problem setup can be generally described as a flavor of 'transfer learning': can the model improve its task related to unsupervised learning by leveraging the availability of labels that are primarily associated to the supervised learning task?

This was investigated in two different kind of experiments: (a) we benchmarked the quality of the representations directly via the betaVAE score as a function of available labels (section \ref{sec:disentanglment}). In this case the added $\pi$ layer can be interpreted as an additional loss term directly reflecting the betaVEA score. 

And (b) we used the VAE as an anomaly detector (see sections \ref{sec:mnist:anomaly}, \ref{sec:fashionmnist:anomaly} and \ref{sec:ucihar}). In this case our approach can be viewed as feature engineering: usually labels incorporate domain knowledge of some very specific, yet important, property of the data set. Thereby our method can guide the $\pi$ layer towards an extractor for those very specific high-level features. In this experiment, we contrasted the semi-supervised model to an equivalent purely unsupervised VAE by removing the $\pi$ layer from the network and the loss function (this corresponds to the left and right most panels of Figure \ref{fig:model_architectures}). We then compared its anomaly detection performance with the anomaly detection performance of our model trained either on a portion or all normal datapoints labeled. 
 
The term 'semi-unsupervised learning' is a perfect description of this task - as semi-supervised learning enhances the performance of a supervised task by using unlabeled data, 'semi-unsupervised' learning would enhance the performance of an unsupervised task by using labeled data. The only other mention of the term was used to describe experiments \cite{willets2019semiunsupervised} \cite{willets2018semiunsupervised} on some other variations of the classic VAE \cite{kingma2013vae}. This unsupervised task was however quite different, in which the objecting was to cluster unlabeled datapoints and subsequently classify them using one-shot-learning.
 
\section{Experiments}
The networks used are described in detail in appendix \ref{app:architectures}.  Each semi-supervised model $SS$ (our architecture as described above in \ref{sec:model_architecture}
), was contrasted with two sibling networks: (a) An equivalent supervised network $ES$, corresponding only to the encoder plus the $\pi$ layer of the $SS$, being trained only on the cross-entropy loss. And (b) an equivalent unsupervised network $EU$, which is identical to our $SS$ architecture, but with the $\pi$ layer removed. Throughout this section, we will refer to our models using the abbreviations as shown in Table \ref{tab:abbreviations}. The error bars were generated by re-running each scenario at least 10 times (unless specified otherwise).

\begin{table}[h]
\begin{center}
\caption{Model abbreviations. Generally speaking the equivalent models $ES$ and $EU$ are derived from $SS$ by either removing the resampling step and decoder (thus making it supervised, $ES$) or by removing the $\pi$ layer (thus making it fully unsupervised, $EU$).}
\label{tab:abbreviations}
\begin{tabular}{ l | l | l | l }

   & Dense & Convolutional & Recurrent \\
   \hline
   Semi-Supervised (ours) & $SS_D$ & $SS_{CNN}$ & $SS_{RNN}$ \\
   \hline
  Equivalent Supervised & $ES_{D}$ & $ES_{CNN}$ & $ES_{RNN}$ \\
  \hline
  Equivalent Unsupervised & $EU_{D}$ & $EU_{CNN}$ & $EU_{RNN}$ \\
\end{tabular}
\end{center}
\end{table}

\subsection{MNIST}
\label{sec:mnist}

For almost any type of image classifier, the natural place to begin any benchmarking is by using the MNIST \cite{lecun2010mnist} dataset, a well known dataset containing $70,000$ ($60,000$ training and $10,000$ testing) grey-scale images of hand written digits. Given the versatility of the model, we conducted the following three benchmarks.

\subsubsection{Semi-Supervised performance}
\label{sec:mnist:classification}

The first task is semi-supervised learning, the area within which the model was designed to bring the most benefit. To create a semi-supervised dataset, we simply discard a certain percentage of the labels, but not the images themselves. This means an equivalent supervised model will only be able to train on the sample of the dataset which is labeled, whereas the semi-supervised model will benefit from all the additional unlabeled samples.

The semi-supervised model was trained for $10$ epochs, whilst the supervised equivalent was trained for $20$ epochs, such that the comparison is made between fully converged models. For comparison, $10$ different labeled subsets of the dataset were taken, with each model trained on identical data, the results of which are displayed in Table \ref{tab:ss_mnist}.

\begin{table}[h]
\begin{center}
\caption{Semi-Supervised MNIST classification results}
\label{tab:ss_mnist}
\begin{tabular}{ l | l | l | l  | l }

	 \multirow{2}{*}{model} & \multicolumn{2}{c |}{$100$ labels} & \multicolumn{2}{c}{$1,000$ labels} \\
	 \cline{2-5}
   & accuracy  & log loss & accuracy & log loss \\
	\hline
  $SS_D$ & $0.808 \pm 0.006$ & $0.80 \pm 0.01$  & $0.9346 \pm 0.0009$ & $\mathbf{0.215 \pm 0.002}$ \\
  \hline
  $ES_{D}$ & $0.763 \pm 0.003$ & $0.75 \pm 0.01$ & $0.902 \pm 0.006$ &  $0.41 \pm 0.03$ \\
  \hline
  $SS_{CNN}$ & $\mathbf{0.811 \pm 0.006}$ & $0.86 \pm 0.03$ & $\mathbf{0.945 \pm 0.001}$ & $0.218 \pm 0.06$  \\
  \hline
  $ES_{CNN}$ & $0.765 \pm 0.003$ & $\mathbf{0.744 \pm 0.009}$ & $0.89 \pm 0.02$ & $0.6 \pm 0.2$ \\
\end{tabular}
\end{center}
\end{table}

Unsurprisingly, both variants of the semi-supervised model outperform their purely supervised equivalents for both sets of labels. For the CNN variant, there was a clear increase in the accuracy of the classification for both $100$ and $1,000$ labels as it scored  $4.6\%$ and $5.5\%$ higher than the supervised equivalent respectively. 

Interestingly, on $100$ labels the log loss was lower for the supervised model than for the semi-supervised model for both the dense and CNN variants; however when trained on $1,000$ labels this trend was reversed. A possible interpretation of this could be that the supervised model is trained on a much smaller dataset and hence starts to overfit, producing predictions with a higher certainty. This might be beneficial for the log loss since the supervised model has predictive power nonetheless as is corroborated by the accuracy score.

\subsubsection{Decoder as a regularizer}
The next test for the model is in the purely supervised domain, testing the hypothesis that the decoder acts as a regularizer and assists the model in finding better representations of the dataset. For this test, both models were trained on the full ($60,000$ sample) training dataset until converged, with the results displayed in Table \ref{tab:reg_mnist}.

\begin{table}[h]
\begin{center}
\caption{Supervised MNIST classification results}
\label{tab:reg_mnist}
\begin{tabular}{ l | l  | l }

  model & accuracy & log loss \\
  \hline
  $SS_D$ & $0.9855 \pm 0.0003$ & $0.062 \pm 0.001$  \\
  \hline
  $ES_{D}$ & $0.9814 \pm 0.0003$ & $0.131 \pm 0.003$ \\
  \hline
  $SS_{CNN}$ & $\mathbf{0.9916 \pm 0.0003}$ & $\mathbf{0.036 \pm 0.002}$  \\
  \hline
  $ES_{CNN}$ & $0.9904 \pm 0.0009$ & $0.055 \pm 0.009$  \\
\end{tabular}
\end{center}
\end{table}

For both the dense and CNN cases the accuracy scores were very similar and, in the case of the CNN model, the error bars are almost overlapping. The log loss of the semi-supervised model however was significantly lower, in particular for the dense model there was a $50\%$ reduction when compared with that of the supervised model. This strongly suggests that the addition of the reconstruction task introduces a much higher confidence in the classifications of the semi-supervised model.

\subsubsection{Semi-Unsupervised Learning}
\label{sec:mnist:anomaly}
The purpose of the semi-unsupervised task was to verify that the introduction of labels can improve the performance of an anomaly detection task. The results presented in Table \ref{tab:ad_mnist} were obtained by training the model on $9$ of the $10$ classes (designated 'normal' using the terminology of anomaly detection) in the MNIST dataset and inferring on all classes, essentially declaring the left out class to be 'anomalous' data. The anomaly score returned by the models is the log reconstruction probability, which is expected to be higher for the anomalous classes. The performance of the scores are evaluated using the AUC (area under curve, also ROC) score.

\begin{table}[h]
\begin{center}
\caption{Label assisted results of anomaly detection on MNIST}
\label{tab:ad_mnist}
\begin{tabular}{ l | l | l }
  anomalous class & AUC of $EU_{D}$  & AUC of $SS_D$ \\
   \hline
  $0$ & $0.949 \pm 0.001$  & $\mathbf{0.969 \pm 0.001}$ \\
  \hline  
  $1$ & $\mathbf{0.47 \pm 0.03}$ & $0.095 \pm 0.006$ \\
  \hline
  $2$ & $0.9610 \pm 0.0009$  & $\mathbf{0.9719 \pm 0.0004}$ \\
  \hline
  $3$ & $0.848 \pm 0.003$  & $\mathbf{0.902 \pm 0.002}$ \\
  \hline
  $4$ & $0.708 \pm 0.004$ & $\mathbf{0.751 \pm 0.003}$ \\
  \hline
  $5$ & $0.860 \pm 0.003$ & $\mathbf{0.894 \pm 0.001}$ \\
  \hline
  $6$ & $0.9295 \pm 0.0008$ & $\mathbf{0.960 \pm 0.002}$ \\
  \hline
  $7$ & $0.669 \pm 0.002$ & $\mathbf{0.755 \pm 0.005}$ \\
  \hline
  $8$ & $0.891 \pm 0.004$ & $\mathbf{0.922 \pm 0.002}$ \\
  \hline
 $9$ & $0.62 \pm 0.02$  & $\mathbf{0.677 \pm 0.005}$ \\
\end{tabular}
\end{center}
\end{table}

With the exception of the digit $1$,  there was a considerable improvement for each anomalous class, with an average improvement of $4.1\%$ over the purely unsupervised model. It demonstrates that the addition of labels aids the model in learning a better representation of normal data. Again, as hypothesized, this is most likely due to the additional label information assisting the model in identifying the category as important high level feature.

The very poor performance of using the digit $1$ as the anomaly class could also be evidence to support this. Given the similarities between the digits $1$ and $7$, it is likely that the dense representation found by the model for the digit $7$ was also sufficient for reconstructing the $1$'s, especially given that the shape of the digit $1$ is most often also found within the digit $7$. This would also explain why there was not a similar drop in performance when considering $7$ as the anomaly class. Considering clustered embeddings, the dense representation of the digit $1$ would not be enough to properly reconstruct a $7$, leading to a higher anomaly score

Based on these results, a further experiment was run to assess the effect of the amount of available labels. Classes '7' and '9' were chosen as they achieved the largest improvement over the unsupervised equivalent and the results are displayed in Table \ref{tab:ad_mnist2}.

\begin{table}[h]
\begin{center}
\caption{Anomaly detection AUC w.r.t. label availability}
\label{tab:ad_mnist2}
\begin{tabular}{ l | l | l }
  label $\%$ & anomaly class $7$ & anomaly class $9$ \\
  \hline
  $1\%$  & $0.751 \pm 0.006$ & $0.670 \pm 0.004$ \\
    \hline
  $10\%$  & $0.736 \pm 0.009$ & $0.679 \pm 0.003$  \\
    \hline
  $25\%$  & $ 0.744 \pm 0.008$ & $0.680 \pm 0.004$  \\
    \hline
  $50\%$  & $0.752 \pm 0.004$ & $0.677 \pm 0.004$ \\
    \hline
  $75\%$  & $0.744 \pm 0.004$ & $0.673 \pm 0.004$  \\
    \hline
  $99\%$  & $0.745 \pm 0.009$ & $0.652 \pm 0.004$  \\
\end{tabular}
\end{center}
\end{table}

For this test, the model was trained by re-sampling the available labels such that the model is trained on an equal amount of labeled and unlabeled data, without making any changes to the testing data. If the label fraction is below $50\%$, this amounts to up-sampling the labeled data, while vice-versa for a label fraction above $50\%$ to downsampling of the labeled data. For both classes, there is almost no difference in the label percentage as the error bars all overlap. Not only is this result somewhat surprising, but it is an advantage of such a model. 

Firstly, it demonstrates that a tiny fraction of labels is all that is required to bring a substantial increase in performance compared to the unsupervised domain, i.e. almost the maximum pay-off can be achieved straight away. 

Secondly, although one would have intuitively expected the performance of the anomaly detector to increase with respect to the number of labels, this is not the case and does not conflict with the hypothesis. Given that the labels are up-sampled during training to balance the learning objective, increasing the percentage of labels in the training set simply increases the diversity of the labeled data rather than the quantity. 

Considering the hypothesis that learning clustered representation of the data assists the model in identifying anomalies, a possible explanation for the lack of improvement in the anomaly detection performance could be reasoned as follows: if a small fraction of labels is enough to push the model to finding such a clustered representation, a larger diversity within the label classes may not provide any further contribution. Perhaps a more diverse representation within the clusters would improve the performance by helping the model to identify anomalous data which lie on the class boundaries.

\subsubsection{Data generation}
\label{sec:mnist_gen}
\begin{figure*}[h]
    \centering
    \begin{subfigure}[t]{0.2\linewidth}
        \centering
        \includegraphics[width=\linewidth]{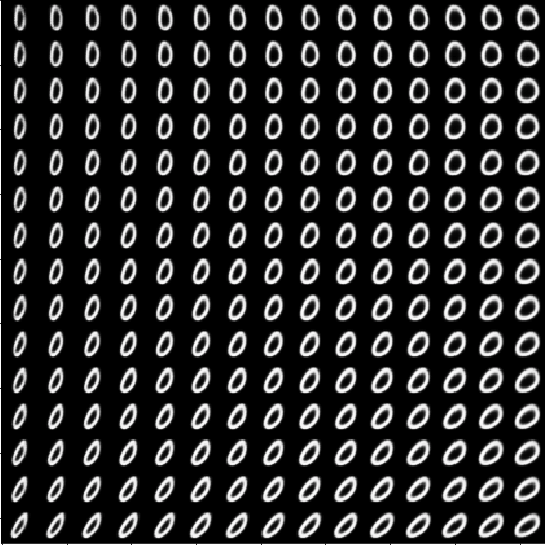}
    \end{subfigure}%
    ~
    \begin{subfigure}[t]{0.2\linewidth}
        \centering
        \includegraphics[width=\linewidth]{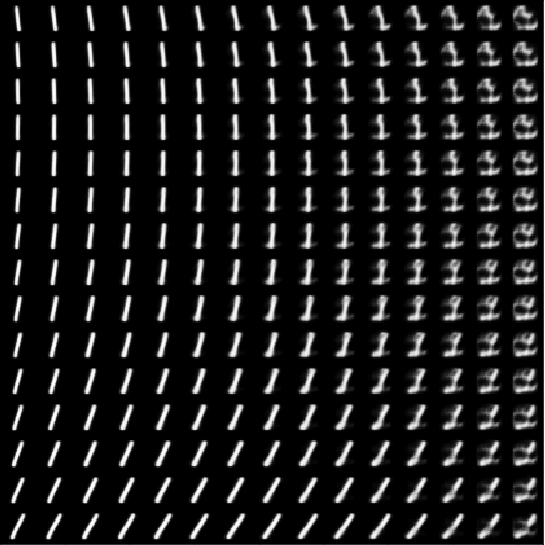}
    \end{subfigure}%
    ~
    \begin{subfigure}[t]{0.2\linewidth}
        \centering
        \includegraphics[width=\linewidth]{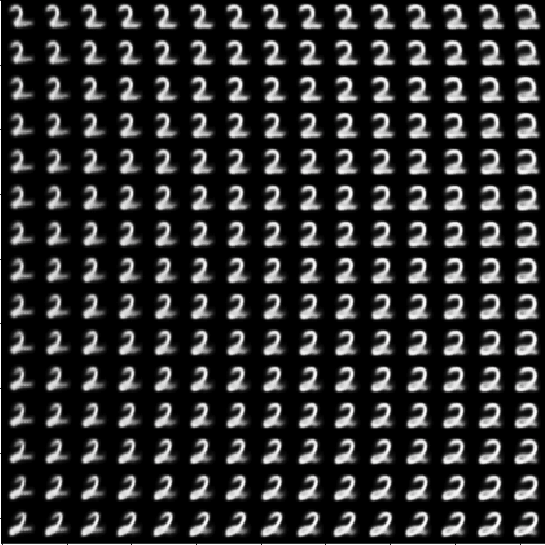}
    \end{subfigure}%
    ~
    \begin{subfigure}[t]{0.2\linewidth}
        \centering
        \includegraphics[width=\linewidth]{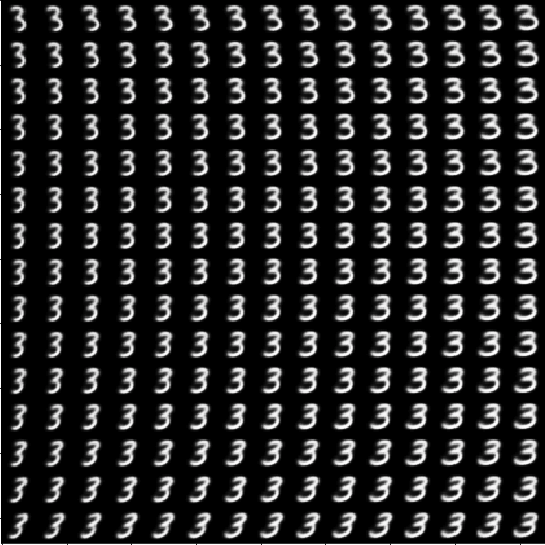}
    \end{subfigure}%
    ~
    \begin{subfigure}[t]{0.2\linewidth}
        \centering
        \includegraphics[width=\linewidth]{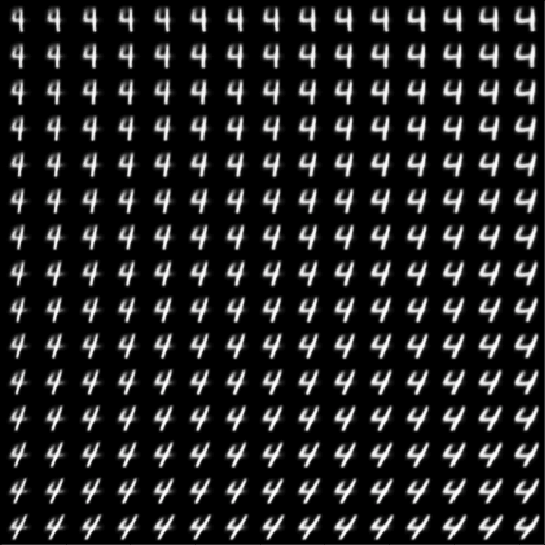}
    \end{subfigure}%
    
    \begin{subfigure}[t]{0.2\linewidth}
        \centering
        \includegraphics[width=\linewidth]{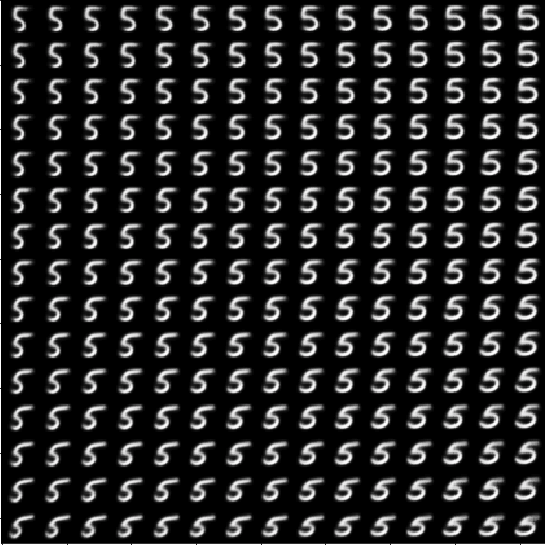}
    \end{subfigure}%
    ~
    \begin{subfigure}[t]{0.2\linewidth}
        \centering
        \includegraphics[width=\linewidth]{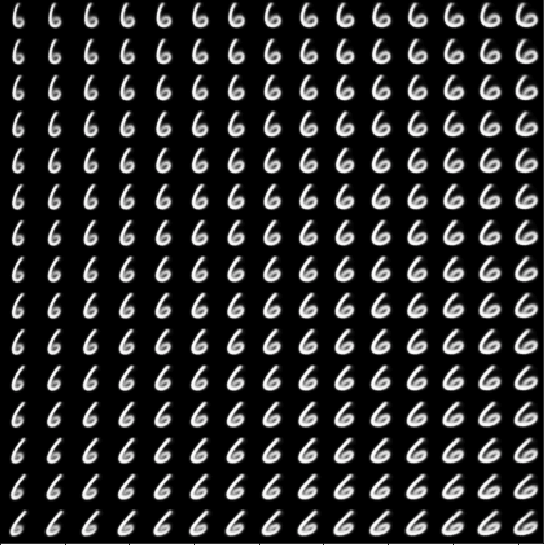}
    \end{subfigure}%
    ~
    \begin{subfigure}[t]{0.2\linewidth}
        \centering
        \includegraphics[width=\linewidth]{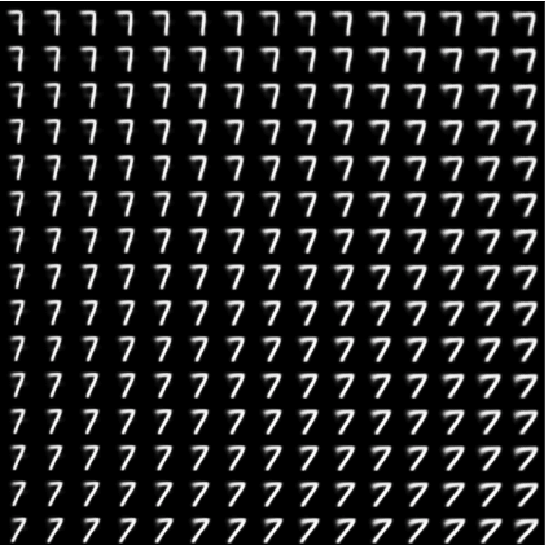}
    \end{subfigure}%
    ~
    \begin{subfigure}[t]{0.2\linewidth}
        \centering
        \includegraphics[width=\linewidth]{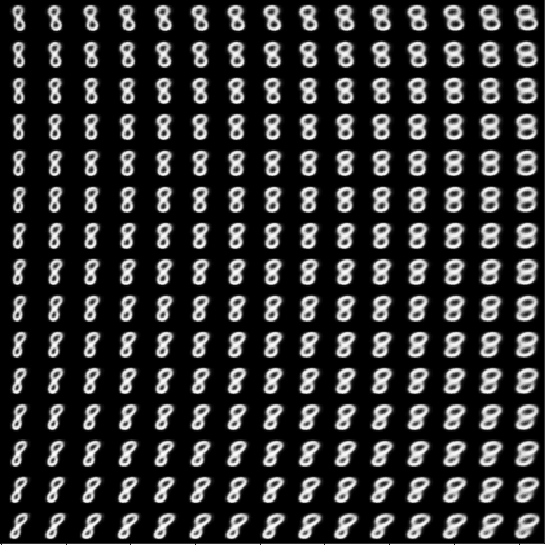}
    \end{subfigure}%
    ~
    \begin{subfigure}[t]{0.2\linewidth}
        \centering
        \includegraphics[width=\linewidth]{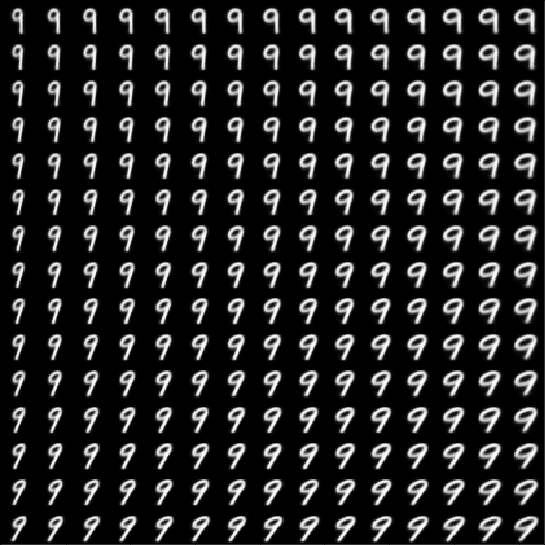}
    \end{subfigure}%
    \caption{Digits generated by our Semi-Supervised model when trained on MNIST}
    \label{fig:mnist_gen}
\end{figure*}

The decoder part of a VAE samples the latent layer and attempts to reconstruct the original input. An advantage of the semi-supervised variant is that the decoder can be used as a generative model by providing both the target label and by sampling from the latent layer. Given that the prior distribution of the latent layer is Gaussian, we can simply sample from a normal distribution to feed as an input to the decoder. Depending on where we sample the normal distribution, the digit which is generated will be a representation of a different region of the training data. In other words, we can separate both class and style. This is demonstrated in Figure \ref{fig:mnist_gen}, which illustrates the range of styles for each digit the model has learned.

\subsection{Fashion-MNIST}
\label{sec:fashionmnist}
Fashion-MNIST \cite{xiao2017fashionmnist} is an image dataset published by Zalando. It is designed as a drop-in replacement of MNIST, i.e. exactly like the original MNIST dataset, Fashion-MNIST contains of $60,000$ training (and $10,000$ testing) $28 \times 28$ grey-scale images with a total of $10$ classes. However, there is much more variability within a given Fashion-MNIST class than there is within a given MNIST class. For instance, the individual samples of the class 'Ankle Boot' vary much more wildly than any digit in the MNIST dataset. As a consequence, Fashion-MNIST is a much more challenging dataset than MNIST, and hence serves as more realistic proxy to evaluate  model performance; especially as most modern image classification models can almost perfectly solve MNIST. At the same time, Fashion-MNIST preserves the big advantage of MNIST: It is still a small dataset that allows rapid training and experimentation when researching new models.

\subsubsection{Semi-Supervised performance}
\label{sec:fashionmnist:classification}
The set-up for the semi-supervised classification task is the same as with the original MNIST dataset. That is, only the labels for a subset of all samples the dataset are retained. The models $ES_{D}$ and $ES_{CNN}$ are then only trained on that subset, whilst the $SS_D$ and $SS_{CNN}$ models are trained on all the samples, but only make use of the labels of the subset. The other samples are treated as unlabeled. The results are shown in Table \ref{tab:ss_fmnist}.

\begin{table}[h]
\begin{center}
\caption{Semi-Supervised Fashion-MNIST classification results}
\label{tab:ss_fmnist}
\begin{tabular}{ l | l | l | l | l}

	\multirow{2}{*}{model} & \multicolumn{2}{c |}{$100$ labels} & \multicolumn{2}{c}{$1,000$ labels} \\
	 \cline{2-5}
   & accuracy  & log loss & accuracy & log loss \\
	\hline
  $SS_D$& $0.703 \pm 0.007$ & $1.3 \pm 0.2$ & $0.812 \pm 0.004$ & $1.19 \pm 0.07$\\
  \hline
  $ES_{D}$ & $0.668 \pm 0.008$ & $1.4 \pm 0.1$ & $0.766 \pm 0.008$ & $1.6 \pm 0.1$ \\
  \hline
  $SS_{CNN}$ & $\mathbf{0.724 \pm 0.008}$ & $\mathbf{1.19 \pm 0.05}$ & $\mathbf{0.836 \pm 0.001}$ & $\mathbf{0.83 \pm 0.01}$  \\
  \hline
  $ES_{CNN}$ & $0.66 \pm 0.01$ & $1.5 \pm 0.1$ & $0.803 \pm 0.004$ & $1.20 \pm 0.05$ \\

\end{tabular}
\end{center}
\end{table}

The difference in the accuracy scores between the semi-supervised and their supervised counterparts is almost identical as with the original MNIST. Unlike the original MNIST however, there is a clear improvement in the log loss from each of the semi-supervised models.

\subsubsection{Decoder as a regularizer}
The results of training our model and an equivalent supervised model on the full dataset with every datapoint labeled (see section \ref{sec:decoder_regularizer}) are displayed in Table \ref{tab:ad_fmnist}.

\begin{table}[h]
\begin{center}
\caption{Supervised Fashion-MNIST classification results}
\label{tab:reg_fmnist}
\begin{tabular}{ l | l | l }
  model & mean accuracy & log loss \\
   \hline
  $SS_D$ & $0.877 \pm 0.004$ & $0.4 \pm 0.02$ \\
  \hline
  $ES_{D}$ & $0.79 \pm 0.01$ & $4.8 \pm 0.4$  \\
  \hline
  $SS_{CNN}$ & $\mathbf{0.925 \pm 0.001}$ & $0.33 \pm 0.025$  \\
  \hline
  $ES_{CNN}$ & $0.922 \pm 0.0015$ & $0.33 \pm 0.015$
\end{tabular}
\end{center}
\end{table}

As can be seen, the fully connected (dense) model profits massively from using the decoder as a regularizer. For the CNN model, the effect is less pronounced. However, the best accuracy is still achieved by $SS_{CNN}$.

\subsubsection{Semi-unsupervised learning}
\label{sec:fashionmnist:anomaly}
The anomaly detection task was set up in the same way as with the original MNIST dataset. One of the 10 classes was designated anomalous, the others were declared normal. The model is trained on a subset of the samples from the remaining nine classes. The held out samples from these nine classes and the samples from the anomalous class are used as a validation set, with the performance of the results evaluated using the AUC score. The results are summarized in Table \ref{tab:ad_fmnist}. The error bounds were obtained by rerunning every experiment five times. 

\begin{table}[h]
\begin{center}
\caption{Label assisted results of anomaly detection on Fashion-MNIST}
\label{tab:ad_fmnist}
\begin{tabular}{ l | l | l }
  anomalous class & AUC of $SS_D$ & AUC of $EU_{D}$ \\
   \hline
  T-shirt/top & $\mathbf{0.712 \pm 0.002}$ & $0.695 \pm 0.002$ \\
  \hline
  Trouser & $\mathbf{0.371 \pm 0.005}$ & $0.347 \pm 0.002$ \\
  \hline
  Pullover & $\mathbf{0.826 \pm 0.005}$ & $0.8 \pm 0.001$ \\
  \hline
  Dress & $\mathbf{0.462 \pm 0.003} $ & $0.416 \pm 0.004$ \\
  \hline
  Coat & $\mathbf{0.719 \pm 0.007}$ & $0.681 \pm 0.001$ \\
  \hline
  Sandal & $\mathbf{0.413 \pm 0.007}$ & $0.34 \pm 0.002$ \\
  \hline
  Shirt & $0.76 \pm 0.002$ & $0.762 \pm 0.001$ \\
  \hline
  Sneaker & $0.191 \pm 0.006$ & $\mathbf{0.21 \pm 0.002}$ \\
  \hline
  Bag & $\mathbf{0.899 \pm 0.009}$ & $0.871 \pm 0.004$ \\
  \hline
  Ankle Boot & $0.525 \pm 0.006$ & $\mathbf{0.547 \pm 0.004}$ \\
\end{tabular}
\end{center}
\end{table}

$SS_D$ performs better than $EU_{D}$ for most anomalous classes. The only exceptions are the 'Shirt class', where there is a tie between both models within the denoted error bar, while $EU_{D}$ wins for the 'Sneaker class' and the 'Ankle Boot class'; although the performance for the 'Sneaker class' for both models is extremely bad.

In general, the model performance varies drastically depending on the anomalous class. While they perform very well for classes like 'Pullover' or 'Bag', they struggle with the 'Sneaker', 'Ankle Boot', 'Trouser' and 'Dress' classes. This can be understood by looking at a samples of each class (see Figure \ref{fig:fashion_mnist}). In particular the 'Sneaker' and 'Ankle Boot' classes bear a resemblances, which explains why an anomaly detector trained on the 'Sneaker' class has hard time to flag the 'Ankle Boot' samples as anomalous (and vice versa). The same is true, though to a lesser degree, for the 'Trouser' and 'Dress' classes.

%TODO
%Run experiments for CNN as well
%
%TODO
%why is the model able to differentiate pullovers from coats so well, but struggles with trousers and dresses? looking at the reconstructions, it seems that the reason is that a very narrow trouser is degenerated into a narrow strip, and the same is true for the dress. however, this probably requires an additional analysis. if this hypothesis is correct, the cnn model should perform much better in that regard, as its reconstructions dont seem to degenerate

\subsubsection{Data Generation}
\begin{figure*}[h]
    \centering
    \begin{subfigure}[t]{0.2\linewidth}
        \centering
        \includegraphics[width=\linewidth]{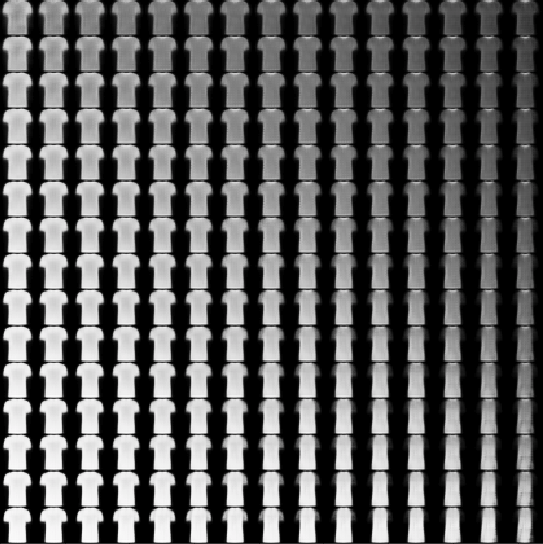}
    \end{subfigure}%
    ~
    \begin{subfigure}[t]{0.2\linewidth}
        \centering
        \includegraphics[width=\linewidth]{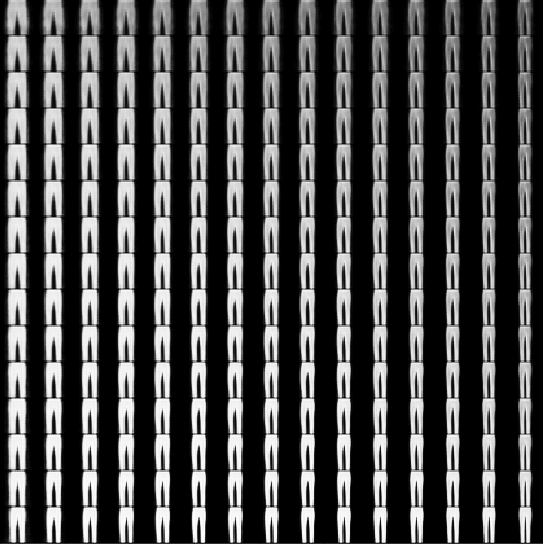}
    \end{subfigure}%
    ~
    \begin{subfigure}[t]{0.2\linewidth}
        \centering
        \includegraphics[width=\linewidth]{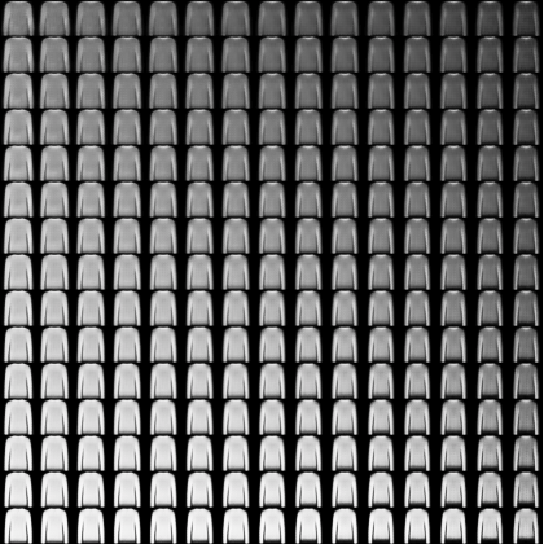}
    \end{subfigure}%
    ~
    \begin{subfigure}[t]{0.2\linewidth}
        \centering
        \includegraphics[width=\linewidth]{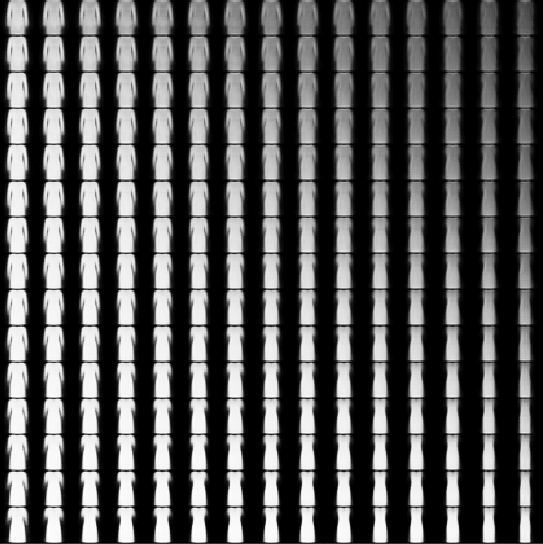}
    \end{subfigure}%
    ~
    \begin{subfigure}[t]{0.2\linewidth}
        \centering
        \includegraphics[width=\linewidth]{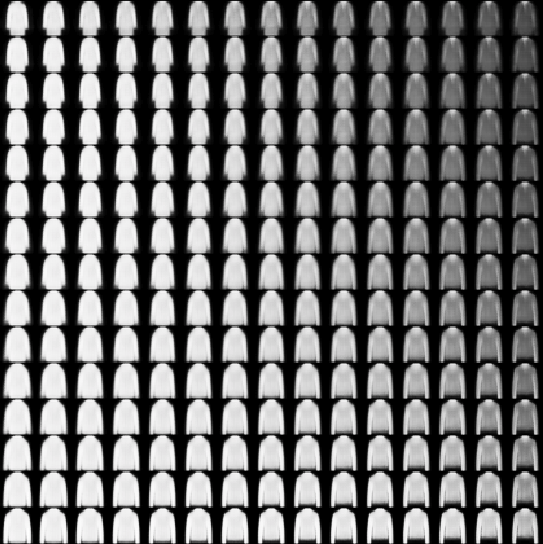}
    \end{subfigure}%
    
    \begin{subfigure}[t]{0.2\linewidth}
        \centering
        \includegraphics[width=\linewidth]{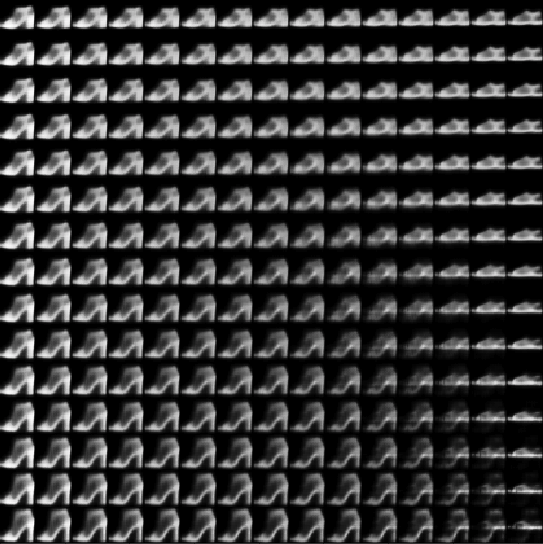}
    \end{subfigure}%
    ~
    \begin{subfigure}[t]{0.2\linewidth}
        \centering
        \includegraphics[width=\linewidth]{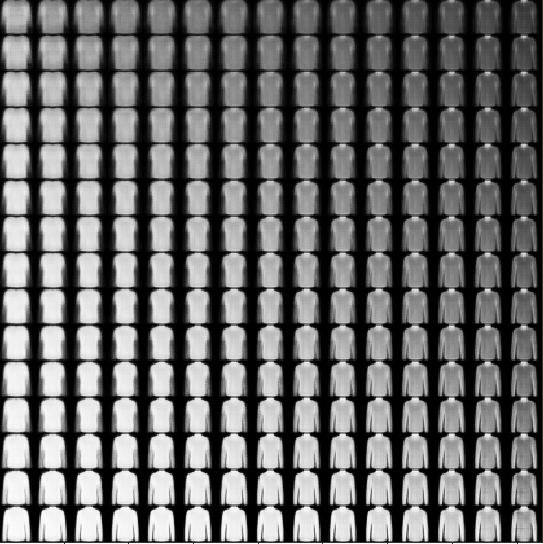}
    \end{subfigure}%
    ~
    \begin{subfigure}[t]{0.2\linewidth}
        \centering
        \includegraphics[width=\linewidth]{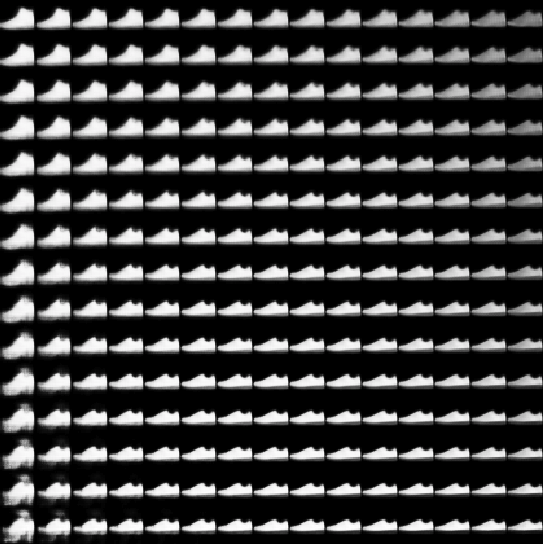}
    \end{subfigure}%
    ~
    \begin{subfigure}[t]{0.2\linewidth}
        \centering
        \includegraphics[width=\linewidth]{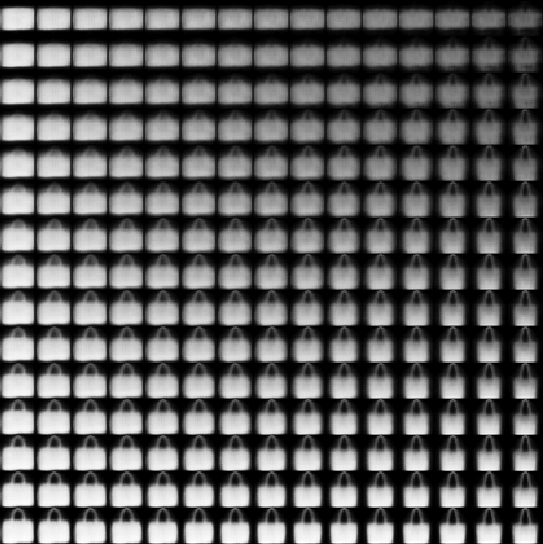}
    \end{subfigure}%
    ~
    \begin{subfigure}[t]{0.2\linewidth}
        \centering
        \includegraphics[width=\linewidth]{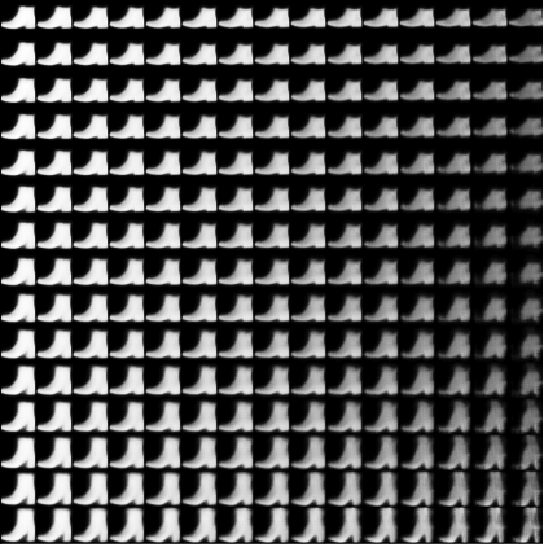}
    \end{subfigure}%
    \caption{Samples generated by our Semi-Supervised model when trained on Fashion-MNIST}
    \label{fig:fmnist_gen}
\end{figure*}

Figure \ref{fig:fmnist_gen} shows an example of generated Fashion-MNIST data, using the same strategy as described in section \ref{sec:mnist_gen}. Once again there is an indication of the styles the model has learned to the embed within the latent space. The style itself is mostly captured in the shape of the class and also the position of highlighted features, for example the position of the straps on the sandals. Comparing the generated data to some of the examples images from Figure \ref{fig:fashion_mnist}, it can be seen that the generated data lacks any of the detailing of the original images. Given that the dataset is more detailed than the digits, but the model architecture is the same, this is likely due to the size of the latent layer and its lack of capacity for storing these additional details.

\subsection{Human activity recognition (UCI-HAR)}
\label{sec:ucihar}
The UCI-HAR dataset \cite{anguita2013ucihar} contains $7,352$ samples, consisting of gyroscopic data recorded from humans, labeled with one of the following six activities: walking, walking upstairs, walking downstairs, sitting, standing and laying. 

The testing conducted on the UCI-HAR dataset was not as thorough as with MNIST and Fashion-MNIST, and we performed only one run per test scenario (hence no error bars are given). The classification results are presented in Table \ref{tab:uci_har} and compare the performance of the semi-supervised model to its supervised equivalent.

\begin{table}[h]
\begin{center}
\caption{UCI-HAR classification results}
\label{tab:uci_har}
\begin{tabular}{ l | l | l | l | l | l | l }

	\multirow{2}{*}{model} & \multicolumn{2}{c |}{$100$ labels} & \multicolumn{2}{c | }{$1,000$ labels} & \multicolumn{2}{c | }{$7,352$ labels}\\
	 \cline{2-7}
   & accuracy  & log loss & accuracy & log loss  & accuracy & log loss \\
   \hline
 	$SS_R$ & $\mathbf{0.630}$ & $4.000$ & $\mathbf{0.839}$ & $\mathbf{0.503}$ &  $\mathbf{0.917}$ & $\mathbf{0.310}$   \\
	\hline
   $ES_{RNN}$ & $0.381$ & $\mathbf{1.538}$  & $0.690$ & $0.820$  & $0.909$ & $0.428$ \\
\end{tabular}
\end{center}
\end{table}

Again, the semi-supervised model is able to outperform its supervised equivalent for each subset of labels. In this case, the performance gain is particularly high when labels are in short supply, as demonstrated by the $24\%$ improvement over the supervised model for the $100$ labels test.

The performance of the model as an anomaly detector was also briefly evaluated, using 'walking' as the anomaly class. The results, displayed in Table \ref{tab:uci_har_ad}, again show an improvement in performance when providing an anomaly detector with labels.

\begin{table}[h]
\begin{center}
\caption{UCI-HAR anomaly detection results}
\label{tab:uci_har_ad}
\begin{tabular}{ l | l }

	model & AUC score \\
   \hline
 	$SS_{RNN}$ & $\mathbf{0.682}$   \\
	\hline
   $ES_{RNN}$ & $0.641$ \\

\end{tabular}
\end{center}
\end{table}

\section{Disentangled representations}
\label{sec:disentanglment}
One major application of VAEs is to find low dimensional representations of real world data \cite{locatello2019google, tschannen2018}. For this task, disentanglement is considered an important quality metric \cite{tschannen2018, bengio_disentanglement}. Generally speaking, disentanglement attempts to quantify how well a particular framework is able to identify important yet independent generating factors of its dataset. For this, multiple distinct metrics and benchmark data sets have been suggested, yet they have been shown to agree at least on a qualitative level \cite{locatello2019google}. In order to benchmark our semi-supervised model, we chose to benchmark via the betaVAE score \cite{higgins2017} on the Small-NORB data set \cite{smallNORB}. Our scores are calculated based on the latent layer (but not the $\pi$ layer), and are shown in Table \ref{tab:disentangled_results}.

The network architecture is described in the appendix, Table \ref{tab:cnn_arch_small_norb} and is similar to \ref{tab:cnn_arch} with two modifications: (a) the labels of the data set were incorporated using multiple cross-entropy loss terms (and one-hot sigmoid $\pi$ layers) for each of the four dimensions, (b) in this architecture the $\pi$ layers are intended to only function as an additional and sparse loss term on the latent layer, and thus is not forwarded to the decoder (no connection between $\pi$ and 'Merge' in Figure \ref{fig:model_architecture:ssvae}). Latter step is necessary, since otherwise the network could bypass the latent layer using the $\pi$ layer, while maintaining reconstruction quality. 

There are two hyperparameters, $\alpha$, the overall weight of the supervised cross-entropy term (equ. \ref{eq:cl_loss}) and $\beta_{norm}$ the overall weight of the KL-divergence term. Both parameters were kept fixed for all cases of this experiment. They were optimized for maximum disentanglement in the completely unsupervised case. Note that this puts all other cases at an disadvantage, since their optimal hyperparameters presumably differ from the unsupervised case - but this is to emulate a real life scenario in which labeled datapoints are either sparse and thus cannot be used for hyperparameter tuning. We empirically found that good results are achieved when $\alpha$ is selected such that its average contribution to the total loss is about of the same order of magnitude as the reconstruction loss (after training converges). The optimal $\beta_{norm}$ was found to be rather small ($0.25$), in accordance with \cite{locatello2019google}. For the betaVAE score itself, we used a plain logistic regressor. Each datapoint for this classifier was based on the averaged absolute differences between the representations of $2 \times 64$ batches. The classifier was trained and tested with 2048 datapoints each.

The final results are shown in Table \ref{tab:disentangled_results}: We always used about $38,000$ unlabeled datapoints, but augmented training by various amounts of labeled datapoints. As expected our approach can outperform both purely unsupervised and a supervised scenarios significantly. Even a relatively small amount of labeled datapoints ($300$, $\sim 1$\%) seems sufficient. It should be noted, that for few (around $100$) labels there is a small, yet statistically significant, decrease in the betaVAE score. This could be due to the aforementioned fact, that the hyperparameters used where optimized for an unsupervised scenario, yet $100$ labels were not sufficient to offset this disadvantage. 

\begin{table}[h]
  \begin{center}
  \caption{betaVAE score of the representations generated by our semi-supervised model on varying label availability. All results were obtained using the same hyperparameters, which were optimized for the first row.}
  \label{tab:disentangled_results}
  \begin{tabular}{ l | l | l }
  \# unlabeled data & \# labeled data & betaVAE score \\
  \hline
  $38,000$ & $0$ & $82 \pm 0.7$ \\
    \hline
  $38,000$ & $100$ & $76 \pm 1.6$ \\
    \hline
  $38,000$ & $300$ & $83 \pm 1.8$ \\
    \hline
  $38,000$ & $1,000$ & $92 \pm 0.6$ \\
    \hline
  $38,000$ & $3,000$ & $95 \pm 0.6$ \\ 
    \hline
  $0$ & $1,000$ & $87 \pm 0.8$  \\

  \end{tabular}
  \end{center}
  \end{table}

\section{Future work}
It would be interesting to see how much larger networks would benefit from the suggested regularization technique. For instance, the same technique could directly be applied to state-of-the-art computer vision networks like \cite{Xie2017ResNext}. We leave this avenue for future investigation.

While the results of the semi-unpervised learning already look promising, this approach so far makes no use of an additional input that labels could provide: incorporating user feedback by labeling a false-positive and false-negative detection as such, with the goal of suppressing future false-positive/false-negative detections. In principle, our model architecture should allow to incorporate such feedback. 

One way to achieve this would be as follows: Prepare new classes for false-positive and false-negative anomaly detections. In the beginning, there will be no samples in these classes. However, once a sufficient amount of false-positive/false-negative detections accumulated, the model is re-trained. The anomaly score $x$ could then be heuristically adjusted, for instance: 

\begin{equation}
\label{eq:loss_future}
x \rightarrow \frac{1 - p_{fp}}{1 - p_{fn}} \cdot x
\end{equation}

where $p_{fp}$ and $p_{fn}$ are the false-positive and false-negative probabilities, respectively, outputted by the model at inference time.

This treatment would even work when there are no labels available except the false-positive/false-negative assignments.

\section{Conclusions}
With one of the most common issues associated with training machine learning models being the availability of training data, semi-supervised models present the perfect opportunity to take advantage of every available scrap of data. This is a particularly valuable improvement in the supervised domain, given the availability of unlabeled data in comparison to labeled data. 

The value added by a semi-supervised approach is even greater when considering such a model in the context of failure prediction, given that with a purely supervised approach, the dataset would consist entirely of failure events. Depending on the system in question, such a dataset could take decades to collect; an obstacle which would often make a predictive system unobtainable. Taking advantage of the often abundant and easy to produce unlabeled data, this semi-supervised approach  demonstrates the ability to converge towards accurate predictions on only a fraction of the labels.

The versatility of the semi-supervised model proposed in this paper delivers concrete improvements across the entire spectrum of label availability. 

Within the labeled domain, the value added by a semi-supervised approach is even greater when considering such a model in the context of failure prediction. With a purely supervised approach, a training dataset would consist entirely of failure events, requiring the system in question to fail hundreds if not thousands of times to gather a sizeable dataset. Depending on the system in question, collecting such a dataset from scratch could take decades; an obstacle which can often make a predictive system unobtainable. Taking advantage of the often abundant and easy to produce unlabeled data, this semi-supervised approach demonstrates the ability to converge towards an accurate predictive system on only a fraction of the labels. 

In addition to reducing the time to deployment, the model offers further benefits in the supervised learning domain, able to outperform equivalent classifiers due the regularizing effect of the decoder and its associated reconstruction task.

In the purely unsupervised domain, the model achieves identical performance to a VAE, yet demonstrates a huge increase of performance with a tiny fraction of labels. Traditionally, a VAE must find a suitable dense representation of the system it's modelling in the latent layer. With the introduction of labels, the latent activations must not only embed a representation of the system, but also a classification of the system state. Ultimately, this additional information results in improved embedding of the system state, not only enabling classifications, but improving reconstructions. In short, the labels provide the model with a better understanding of the system which it is reconstructing. 

\bibliographystyle{unsrt}  


\appendix
\section{Network architectures and training}
\label{app}
\label{app:architectures}

\subsection{MNIST and Fashion-MNIST: Model 1, FCN}
\label{app:fcc_arch}
We used the raw pixels, nromalized to $(0,1)$ as input, corresponing to size $28 \cdot 28 = 784$ feature vectors. The architecture is shown below in Table \ref{tab:fcc_arch}. All encoder and decoder layers are simply stacked. The latent layers ($\mu$ and $\sigma$ for a gaussian prior) are forked from the last encoder layer, and merged together with the resampled latent layer as input to the decoder. The model is trained with RMSprop \cite{hintonrmsprop} without decay and a momentum parameter of $0.9$. If not explicitly mentioned otherwise, the learning rate was set to $0.0005$.

\begin{table}[H]
  \begin{center}
  \caption{Fully connected network architecture. The input image corresponds to a 28x28 image reshaped into a single feature vector.}
  \label{tab:fcc_arch}
  \begin{tabular}{ l  l | l | c}
   & layer type & dimensions & comments \\
     \hline
      \hline
   \textbf{encoder} & input layer & 784 & \\ 
    \hline
    & fully connected & 1024 & \textit{relu} activation \\
    \hline
    & fully connected & 1024 & \textit{relu} activation \\
    \hline
     \hline
    \textbf{latent layer} & fully connected & 2 & \textit{linear} activation; latent gaussian mean  \\
    \hline
     & fully connected & 2 & \textit{linear} activation; latent gaussian variance  \\
      \hline
     & fully connected & 10 & \textit{softmax} activation;  class prediction \\
     \hline
      \hline
   \textbf{decoder} & fully connected & 1024 & \textit{relu} activation \\ 
    \hline
    & fully connected & 1024 & \textit{relu} activation \\
    \hline
    & output layer & 784 & \textit{sigmoid} activation \\
  \end{tabular}
  \end{center}
  \end{table}

\subsection{MNIST and Fashion-MNIST: Model 2, CNN}
The images were rescaled to $(0,1)$ but not reshaped. Here we used a series of convolutional layers as detailed in Table \ref{app:fcc_arch}. The last dimension in the dimensions column corresponds to the feature dimension, while the first two correspond to the image dimensions. Again, all encoder and decoder layers were simply stacked, whereas the latent and $\pi$ layer were forked from the last encoder layer. The resampled latent layer and $\pi$ layer were concatenated as input for the decoder. The model is trained with RMSprop without decay and a momentum parameter of $0.9$. If not explicitly mentioned otherwise, the learning rate was set to $0.0005$.

\begin{table}[H]
  \begin{center}
  \caption{Convolutional network architecture (CNN). The input shape corresponds to a single greysclae image (x-axis,y-axis, channel). BN is an abbriavtaion for batch normalization layer, ConvCNN for a transposed convolution layer.}
  \label{tab:cnn_arch}
  \begin{tabular}{ l  l | l | c}
   & layer type & dimensions & comments \\
     \hline
      \hline
   \textbf{encoder} & input layer & $(28, 28, 1)$ & \\ 
    \hline
    & CNN & (28, 28, 64) & kernel $3 \times 3$;  stride $1$ \\
    \hline
    & BN & (28, 28, 64) & with \textit{relu} activation \\
    \hline
    & CNN & (28, 28, 64) & kernel $3 \times 3$;  stride  $1$ \\
    \hline
    & BN & (28, 28, 64) & with \textit{relu} activation \\
    \hline
    & CNN & (14, 14, 64) & kernel $3 \times 3$;  stride  $2$ \\
    \hline
    & BN & (14, 14, 64) & with \textit{relu} activation \\
    \hline
    & Flatten & 12544 &  \\
    \hline
    & fully connected & 512 &  \\
    \hline
    & BN & 512 & with \textit{relu} activation \\
     \hline
    & Dropout & 512 & dropout rate $= 0.5$ \\
    \hline
     \hline
    \textbf{latent layer} & fully connected & 2 & \textit{linear} activation; latent gaussian mean  \\
    \hline
     & fully connected & 2 & \textit{linear} activation; latent gaussian variance  \\
      \hline
     & fully connected & 10 & \textit{softmax} activation;  class prediction \\
     \hline
      \hline
   \textbf{decoder} & fully connected & 12544 & \\ 
   \hline
    & BN & 12544 & with \textit{relu} activation \\
    \hline
     & Dropout & 12544 & dropout rate $= 0.5$\\
    \hline
    & Reshape & (14, 14, 64) & \\
     \hline
    & ConvCNN & (14, 14, 64) & kernel $3 \times 3$;  stride  $1$ \\
    \hline
    & BN & (14, 14, 64) & with \textit{relu} activation \\
    \hline
    & ConvCNN & (14, 14, 64) & kernel  $3 \times 3$;  stride  $1$ \\
    \hline
    & BN & (14, 14, 64) & with \textit{relu} activation \\
    \hline
    & ConvCNN & (28, 28, 64) & kernel  $3 \times 3$;  stride  $2$ \\
    \hline
    & BN & (28, 28, 64) & with \textit{relu} activation \\
    \hline
    & ConvCNN (output) & (28, 28, 1) & kernel  $1 \times 1$;  stride  $1$ \\
  \end{tabular}
  \end{center}
  \end{table}

\subsection{UCI-HAR dataset: RNN}
The recurrent VAE flavor was applied on the UCI HAR dataset \cite{anguita2013ucihar}. We used a look-back dimension of $128$ with the full architecture described in the appendix in Table \ref{tab:rnn_arch}. The first dimension in the dimensions column corresponds to the look-back dimension, while the second corresponds to the feature dimension. The latent layers are forked from the last encoder layer, and concatenated prior to the first decoder layer. The model is trained with RMSprop without decay and a momentum parameter of $0.9$. If not explicitly mentioned otherwise, the learning rate was set to $0.001$.

\begin{table}[H]
  \begin{center}
  \caption{RNN network architecture.The input shape corresponds to a single time series with (128 time steps, 6 features).}
  \label{tab:rnn_arch}
  \begin{tabular}{ l  l | l | c}
   & layer type & dimensions & comments \\
     \hline
      \hline
   \textbf{encoder} & input layer & (128, 6) & \\ 
    \hline
    & fully connected along feature dim & (128, 40) & \textit{relu} activation \\
    \hline
    & LSTM & (128, 40) &  \\
    \hline
    & fully connected along feature dim & (128, 30) & \textit{relu} activation \\
    \hline
     \hline
    \textbf{latent layer} & fully connected & (128, 2) & \textit{linear} activation; latent gaussian mean  \\
    \hline
     & fully connected & (128, 2) & \textit{linear} activation; latent gaussian variance  \\
      \hline
     & fully connected & (128, 6) & \textit{softmax} activation;  class prediction \\
     \hline
      \hline
   \textbf{decoder} & fully connected along feature dim & (128, 30) & \textit{relu} activation \\
    \hline
    & LSTM & (128, 40) &  \\
    \hline
     & fully connected along feature dim & (128, 40) & \textit{relu} activation \\
     \hline
     \hline
    \textbf{output} & fully connected along feature dim & (128, 6) & \textit{linear} activation; gaussian mean $\mu$ \\
     \hline
     & fully connected along feature dim & (128, 6) & \textit{softplus} activation; gaussian variance $\sigma$ \\   
  \end{tabular}
  \end{center}
  \end{table}

\subsection{Small-NORB dataset: CNN}
The input images were rescaled to $(0,1)$. Each stero image pair was stacked into a single feature vector of shape $(96,96,2)$. The encoder and decoder are simply stacked and the full architecture is shown in Table \ref{tab:cnn_arch_small_norb}. For each of the four generating factors of this data set (category, elevation, azimuth and lighting) a sperate $\pi$ layer was added after the encoder. The latent layers ($\mu$ and $\sigma$) of a gaussian prior are also added on top of the encoder. By contrast to the other models, only the resampled latent layer is used for the decoder. The model is trained with RMSprop without decay and a momentum parameter of $0.9$. If not explicitly mentioned otherwise, the learning rate was set to $0.001$.

\begin{table}
\begin{center}
\caption{CNN network architecture used for generating represenations of the Small-NORB data set. The input shape corresponds to a single stero image pair (x-axis, y-axis, left/right). BN is an abbriavtaion for batch normalization layer, ConvCNN for a transposed convolution layer.}
\label{tab:cnn_arch_small_norb}
\begin{tabular}{ l  l | l | c}
 & layer type & dimensions & comments \\
   \hline
    \hline
 \textbf{encoder} & input layer & $(96, 96, 2)$ & \\ 
  \hline
  & CNN & (48, 48, 32) & kernel  $7 \times 7$;  stride  $2$ \\
  \hline
  & BN & (48, 48, 32) & with \textit{relu} activation \\

  \hline
  & CNN & (24, 24, 32) & kernel  $7 \times 7$;  stride  $2$ \\
  \hline
  & BN & (24, 24, 32) & with \textit{relu} activation \\

  \hline
  & CNN & (12, 12, 64) & kernel  $7 \times 7$;  stride  $2$ \\
  \hline
  & BN & (12, 12, 64) & with \textit{relu} activation \\

  \hline
  & CNN & (6, 6, 64) & kernel  $7 \times 7$;  stride  $2$ \\
  \hline
  & BN & (6, 6, 64) & with \textit{relu} activation \\

  \hline
  & Flatten & 2304 &  \\
  \hline
  & fully connected & 256 &  \\
  \hline
  & BN & 256 & with \textit{relu} activation \\
   \hline
  & Dropout & 256 & dropout rate $= 0.5$ \\
  \hline

   \hline
  \textbf{latent layer} & fully connected & 32 & \textit{linear} activation; latent gaussian mean  \\
  \hline
   & fully connected & 32 & \textit{linear} activation; latent gaussian variance  \\
    \hline
   & fully connected & 5 & \textit{softmax} activation;  one-hot prediction (category) \\
   \hline
   & fully connected & 9 & \textit{softmax} activation;  one-hot prediction (elevation) \\
   \hline
   & fully connected & 18 & \textit{softmax} activation;  one-hot predictioon (azimuth) \\
   \hline
   & fully connected & 6 & \textit{softmax} activation;  one-hot prediction (lighting) \\
   \hline
    \hline
    
 \textbf{decoder} & fully connected & 2304 & based on resampled gaussian latent layers \\ 
 \hline
  & BN & 2304 & with \textit{relu} activation \\
  \hline
   & Dropout & 2304 & dropout rate $= 0.5$\\
  \hline
  & Reshape & (6, 6, 64) & \\

  \hline
  & ConvCNN & (12, 12, 64) & kernel  $7 \times 7$;  stride  $2$ \\
  \hline
  & BN & (12, 12, 64) & with \textit{relu} activation \\

  \hline
  & ConvCNN & (24, 24, 64) & kernel  $7 \times 7$;  stride  $2$ \\
  \hline
  & BN & (24, 24, 64) & with \textit{relu} activation \\

  \hline
  & ConvCNN & (48, 48, 32) & kernel  $7 \times 7$;  stride  $2$ \\
  \hline
  & BN & (48, 48, 32) & with \textit{relu} activation \\

  \hline
  & ConvCNN & (96, 96, 32) & kernel  $7 \times 7$;  stride  $2$ \\
  \hline
  & BN & (96, 96, 32) & with \textit{relu} activation \\

  \hline
  & ConvCNN (output) & (96, 96, 2) & kernel  $7 \times 7$;  stride  $1$ \\
\end{tabular}
\end{center}
\end{table}

\section{Samples from Fashion-MNIST data set}

\begin{figure}[H]
  \centering
  \includegraphics[width=1.0\linewidth]{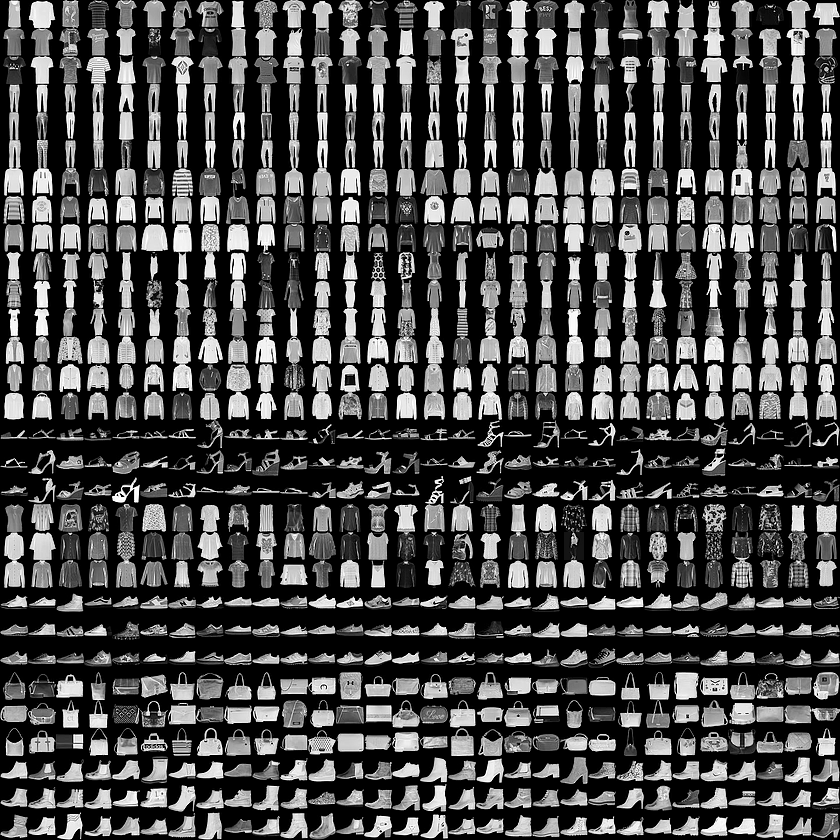}
  \caption{This figure shows some examples from the Fashion-MNIST dataset. Every class always corresponds to three consecutive rows. The classes are (from top to down): T-shirt/top, Trouser, Pullover, Dress, Coat, Sandal, Shirt, Sneaker, Bag and Ankle Boot. }
  \label{fig:fashion_mnist}
\end{figure}

\end{document}